\documentclass[10pt,leqno]{article}
\usepackage[utf8]{inputenc}
\usepackage[T1]{fontenc}
\usepackage{authblk}            
\usepackage{graphicx,amssymb,amsmath,amsthm,csquotes}
\usepackage{xcolor,paralist,hyperref,titlesec,fancyhdr,etoolbox}
\usepackage[margin=1in]{geometry}
\usepackage{lipsum}

\hypersetup{colorlinks=true,linkcolor=black,filecolor=black,urlcolor=black}

\titleformat{\section}[display]{\normalfont\huge\bfseries\centering}{\thesection}{10pt}{\Large}
\titlespacing*{\section}{0pt}{0ex}{0ex}

\usepackage{array} 
\newcolumntype{C}[1]{>{\centering\arraybackslash}m{#1}}

\title{HEMERA: A Human-Explainable Transformer Model for Estimating Lung Cancer Risk using GWAS Data}

\author[1]{Maria Mahbub\thanks{Corresponding author: \texttt{mahbubm@ornl.gov}}}
\author[2,3]{Robert J. Klein}
\author[2,3]{Myvizhi Esai Selvan}
\author[4]{Rowena Yip}
\author[4]{Claudia Henschke}
\author[5]{Providencia Morales}
\author[1]{Ian Goethert}
\author[1]{Olivera Kotevska}
\author[1]{Mayanka Chandra Shekar}
\author[1]{Sean R. Wilkinson}
\author[1]{Eileen McAllister}
\author[5]{Samuel M. Aguayo}
\author[2,3,5]{Zeynep H. G\"um\"u\c{s}}
\author[1]{Ioana Danciu}
\author[6]{VA Million Veteran Program}

\affil[1]{Oak Ridge National Laboratory, Oak Ridge, TN, USA}
\affil[2]{Department of Genetics and Genomics, and Department of AI and Human Health, Icahn School of Medicine at Mount Sinai, New York, NY, USA}
\affil[3]{Marc and Jennifer Lipschultz Precision Immunology Institute, Icahn School of Medicine at Mount Sinai, New York, NY, USA}
\affil[4]{Department of Radiology, Icahn School of Medicine at Mount Sinai, New York, NY, USA}
\affil[5]{Phoenix Veteran Affairs Health Care System, Phoenix, AZ, USA}
\affil[6]{VA Million Veteran Program}

\date{} 

\begin{document}
\maketitle

\begin{abstract}

Lung cancer (LC) is the third most common cancer and the leading cause of cancer deaths in the US. Although smoking is the primary risk factor, the occurrence of LC in never-smokers and familial aggregation studies highlight a genetic component. Genetic biomarkers identified through genome-wide association studies (GWAS) are promising tools for assessing LC risk. We introduce HEMERA (\textbf{H}uman-\textbf{E}xplainable Transformer \textbf{M}odel for \textbf{E}stimating Lung Cancer \textbf{R}isk using GW\textbf{A}S Data), a new framework that applies explainable transformer-based deep learning to GWAS data of single nucleotide polymorphisms (SNPs) for predicting LC risk. Unlike prior approaches, HEMERA directly processes raw genotype data without clinical covariates, introducing additive positional encodings, neural genotype embeddings, and refined variant filtering. A post hoc explainability module based on Layer-wise Integrated Gradients enables attribution of model predictions to specific SNPs, aligning strongly with known LC risk loci. Trained on data from 27,254 Million Veteran Program participants, HEMERA achieved >99\% AUC (area under receiver characteristics) score. These findings support transparent, hypothesis-generating models for personalized LC risk assessment and early intervention.

\end{abstract}


\flushbottom
\maketitle

\thispagestyle{empty}

\section*{Introduction}
LC remains one of the most formidable public health challenges in oncology, ranking as both the deadliest and third most common cancer in the United States \cite{seer_common_cancers}.
While tobacco smoking is the principal known risk factor, a significant number of cases occur in never-smokers.
Twin studies have suggested an important role for inherited genetic susceptibility in LC development that extends beyond traditional environmental exposures \cite{mucci2016familial, cheng2021lung}.
Early detection is vital for better prognosis, yet LC is frequently diagnosed at advanced stages, making it particularly lethal \cite{geddes1979natural, birring2005symptoms}.
These challenges, compounded by the resource-intensive nature of widespread screening programs \cite{martini2021ongoing}, highlight the critical need for more precise risk prediction methodologies that can drive individualized care. 
Existing screening protocols rely primarily on smoking history and age, overlooking genetically predisposed individuals who do not meet conventional eligibility criteria \cite{uspstf_lung_screening}.
These challenges highlight the critical need for more precise risk prediction methodologies. 

The rise of genomic medicine has opened new frontiers in LC risk prediction.
Genome-wide association studies (GWAS) have uncovered numerous susceptibility loci linked to LC risk across diverse populations \cite{wang2016meta, bosse2018decade}.
Genetic variants including single nucleotide polymorphisms (SNPs), copy number variations (CNVs), and rare pathogenic mutations have demonstrated significant associations with LC susceptibility \cite{tian2024single, de2020inherited, khan2024predisposing, selvan2020inherited, esai2019rare}, thereby showing promising predictive value to enable more targeted and cost-effective screening strategies, potentially revolutionizing early detection and improving patient outcomes.
However, integrating these complex, high-dimensional data into clinically actionable precision medicine risk models remains a significant challenge, requiring advanced computational frameworks that can capture non-linear interactions, quantify uncertainty, and offer explainability to support clinical decision-making.
Twin and array-based studies estimate LC heritability at approximately 8–20\% \cite{mucci2016familial, dai2017estimation, jiang2019shared, byun2021shared, sampson2015analysis}, and GWAS efforts have identified around 45 genomic loci associated with risk—particularly influencing susceptibility across different histological subtypes, ancestries, and smoking statuses \cite{long2022functional}.
However, uncovering the biological mechanisms behind these associations remains challenging due to correlation between variants (linkage disequilibrium), noncoding variant effects, and context-specific gene regulation \cite{gorman2024multi, yang2023integrating}.

A substantial body of research has focused on leveraging genetic data for cancer and other disease risk prediction, particularly through the use of computational tools such as polygenic risk scores (PRS). PRS approaches aggregate the effects of multiple SNPs identified from GWAS to estimate individual-level disease risk \cite{chatterjee2013projecting}.
While effective in quantifying risk, these models typically rely on linear assumptions and often lack the capacity to capture epistatic interactions and context-specific regulatory effects, limiting their predictive accuracy and explainability in complex diseases like LC \cite{gorman2024multi, torkamani2018personal, esai2019rare, klein2022polygenic}.
To address these shortcomings, machine learning approaches -- including random forests, support vector machines, and neural networks -- have been explored for genetic risk prediction \cite{kruppa2012risk, yang2022machine, sigala2023machine}.
These models offer increased flexibility and can model complex, non-linear relationships among high-dimensional genomic features \cite{azodi2020opening, libbrecht2015machine}.
Among recent advances, transformer-based architectures have demonstrated superior performance across a range of sequence modeling tasks, including those in computational biology and genomics \cite{ji2021dnabert, zhang2023applications, elmes2022snvformer, collienne2025accuracy, lee2024genotype}.
Their attention mechanisms enable efficient handling of long-range dependencies and complex interactions--capabilities that are particularly suited to the sparse and structured nature of GWAS data.
However, despite these advantages, such models are often regarded as ``black boxes'' due to their limited transparency, contributing to challenges for clinical adoption where explainability and mechanistic inference are essential \cite{azodi2020opening, sadeghi2024review}.

Transformer-based models hold potential for applications in genomic medicine, but their use for cancer risk prediction -- and specifically for LC risk prediction using GWAS data -- remains largely unexplored.
Recent efforts such as Genetformer \cite{thaalbi2025genetformer}, Gene Swin Transformer \cite{wang2025gene}, SNVformer \cite{elmes2022snvformer}, and transformer-powered graph representation learning \cite{su2025interpretable} demonstrate the utility of attention-based models in predicting various cancer risks by capturing complex, non-linear patterns in high-dimensional omics datasets.
Very few studies have directly focused on LC risk prediction using GWAS variant data with transformer models.
GPformer \cite{wu2023transformer} integrates knowledge-guided transformer modules for genomic prediction but has not been applied to disease-specific GWAS datasets.
The GSNDriver framework \cite{bai2024deep} applies transformers to identify LC driver genes from somatic mutation and expression data and achieves strong performance in tumor classification tasks, but it addresses tumor progression rather than inherited risk and does not use germline GWAS data.

\begin{figure}[!ht]
    \centering
    \includegraphics[width=\linewidth]{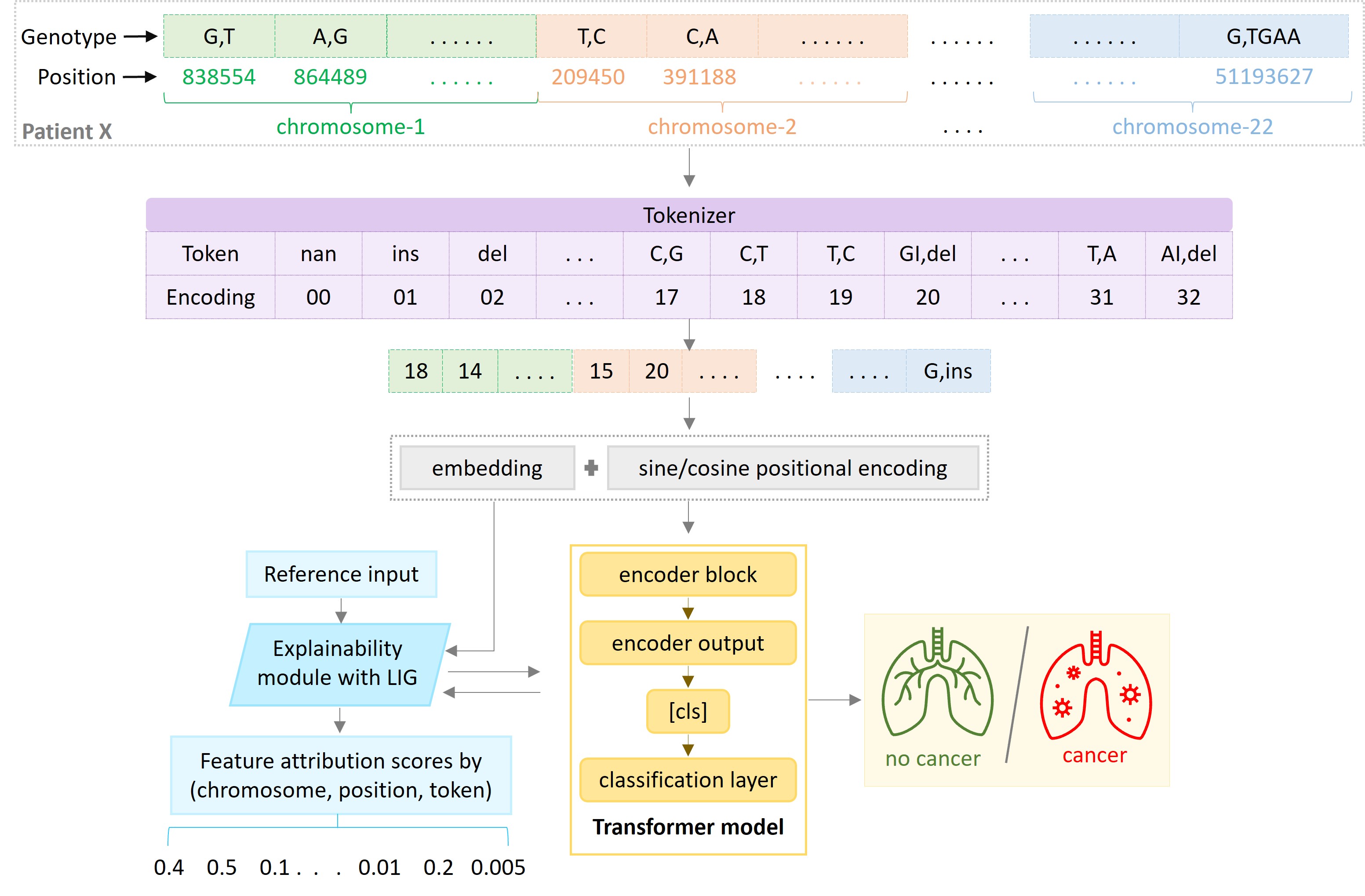}
    \caption{HEMERA: a human-explainable transformer model for estimating lung cancer risk using GWAS data}
    \label{fig:pipeline}
\end{figure}

There is a lack of research on combining GWAS-derived variant data and transformer-based architectures to predict lung cancer susceptibility in an explainable, risk stratified framework \cite{zhang2025lung}.
We aimed to fill this gap by introducing \textbf{HEMERA}: a
\textbf{H}uman-\textbf{E}xplainable Transformer \textbf{M}odel for \textbf{E}stimating Lung Cancer \textbf{R}isk using GW\textbf{A}S Data,
using GWAS data (Fig. \ref{fig:pipeline}).

HEMERA leverages genome-wide association data to deliver both accurate risk stratification and fine-grained feature attribution, bridging the gap between predictive performance and mechanistic understanding of the genetic determinants of LC susceptibility.
The primary contributions of HEMERA are as follows:
\begin{itemize}
\item Unlike prior transformer-based models for cancer risk prediction that combine genetic data with clinical and demographic variables (e.g., age, sex), we isolate inherited genetic variation, allowing us to specifically quantify its predictive contribution without confounding from non-genetic risk factors.
HEMERA departs from this trend and is designed to operate directly on raw genotype data from GWAS, enabling a more principled assessment of inherited genetic risk.
\item HEMERA features a series of critical architectural and methodological innovations over the most relevant prior work \cite{elmes2022snvformer}.
HEMERA introduces additive positional encoding in place of the original concatenation-based scheme, thereby aligning with standard transformer formulations and enhancing training stability and computational efficiency.
It also substitutes conventional one-hot genotype encodings with a neural embedding layer, enabling learnable and semantically rich representations of genetic variants.
The entire data preprocessing pipeline and encoding hyperparameters is re-engineered to incorporate refined variant filtering and dimensionality control to more effectively handle raw genotype calls.
A comprehensive ablation study informs the selection of transformer depth and attention head configurations, ensuring architectural alignment with the complexity of GWAS data.
Finally, HEMERA employs stratified k-fold cross-validation to rigorously assess predictive performance and ensure generalizability across diverse genetic profiles.
\item HEMERA integrates a post hoc explainability module based on Layer-wise Integrated Gradients (LIG), enabling fine-grained attribution of prediction outcomes to specific single nucleotide polymorphisms (SNPs), thereby facilitating biological insight and hypothesis generation.
For validation purposes, the top attributed SNPs are cross-referenced with known LC-associated loci from large-scale GWAS and functional annotation studies.
\end{itemize}

Through its design, HEMERA achieves strong predictive performance while offering explainable, variant-level insights, enabling the identification of putative risk loci associated with LC susceptibility.
By operating solely on raw genotype data, HEMERA highlights the capacity of deep learning models -- specifically transformer-based architectures -- to extract meaningful genomic representations for complex disease phenotypes.
This approach opens new avenues for early detection and enhances our understanding of inherited genetic risk in LC.

\section*{Methods}

We describe in detail the dataset, data preprocessing, model architecture, explainability framework, training configurations, and evaluation metrics below.

\subsection*{Dataset}
\begin{figure}[!htbp]
    \centering
    \includegraphics[width=\linewidth]{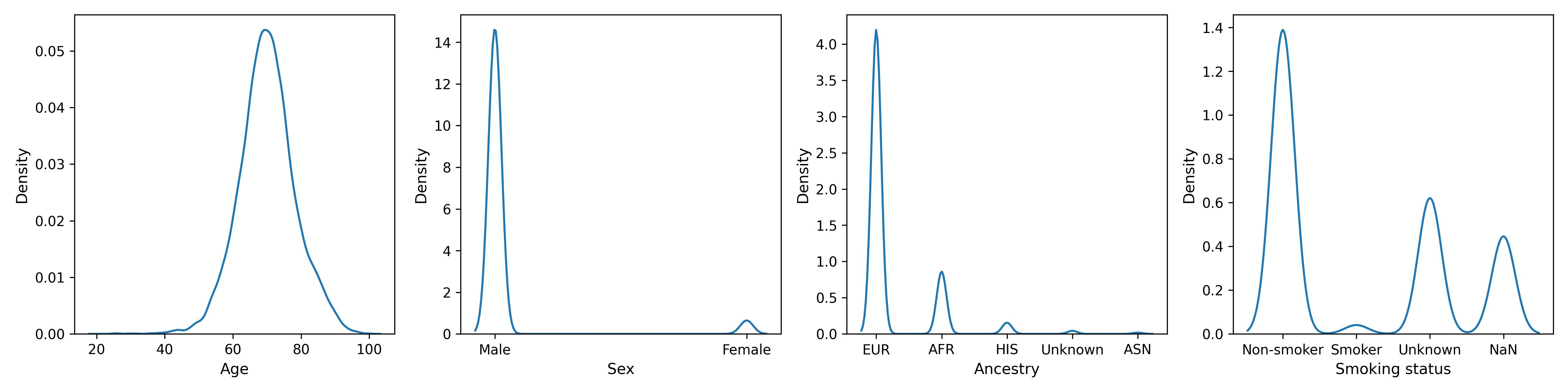}
    \caption{Matched distribution of age, sex, ancestry, and smoking status between cases and controls in the study cohort.}
    \label{fig:data_dist}
\end{figure}

We leveraged array-based genotyping data from participants in the Million Veteran Program (MVP).
Participants who withdrew from the MVP study were excluded from our analysis.
The final cohort included 13,627 participants diagnosed with LC and 13,627 cancer-free controls. 
Controls were matched to cases on key demographic and clinical characteristics, including age, sex, ancestry, and smoking status.
We calculated age at the time of diagnosis for cases and at the time of the last clinical visit for controls.  
Fig. \ref{fig:data_dist} shows the distribution of age, sex, ancestry, and smoking status in the study cohort.
Our GWAS data included 667,955 single nucleotide polymorphisms (SNPs), all of which were directly genotyped without imputation.
All genomic coordinates are referenced to the human genome build GRCh37 (b37).
We performed quality control using PLINKv1.9 \cite{purcell2007plink} removing SNPs with minor allele frequency (MAF) < 0.01. 
Following quality control measures, 378,866 SNPs remained for downstream analysis.
Ethics oversight: This study was approved by the VA Central IRB (MVP061).

\subsection*{Data Preprocessing and Tokenization}
We transformed genomic data into a format suitable for machine learning analysis by considering each participant's SNP sequence as a text sequence, with individual SNPs serving as tokens.
We then prepared the data for modeling following established methods from Elmes et al.\cite{elmes2022snvformer} that consisted of six sequential steps:
(1) major and minor alleles were combined into single tokens depending on the genotype,
(2) insertion (ins) and deletion (del) variants were standardized using dedicated “ins” and “del” tokens to represent any form of insertion and deletion relative to the reference genome,
(3) long nucleotide sequences were compressed using an `I' token to represent all nucleotides following the first nucleotide \cite{cahyawijaya-etal-2022-snp2vec}, which significantly reduced feature space by removing the large number of unique nucleotide sequences,
(4) Unknown or missing SNP calls were encoded as a ‘nan’ token to handle incomplete genotype data,
(5) a special ‘cls’ (classification) token was prepended to each input sequence, with its final hidden representation serving as the aggregate sequence embedding for downstream classification tasks \cite{devlin2019bert},
(6) all tokenized combinations were mapped to integers ranging from 0 to 32, representing the 33 possible token combinations detailed in Table \ref{tab:encode}.
This preprocessing approach enabled efficient representation of genomic variation while maintaining biological interpretability and computational tractability.

\begin{table*}[!htbp]
\caption{Lookup table for encoding SNPs.}
\begin{tabular}{|rrrrrrrr|}
\hline
\multicolumn{8}{|l|}{token : integer}
\\ \hline
\multicolumn{1}{|r|}{nan : 00}
& \multicolumn{1}{r|}{ins : 01}
& \multicolumn{1}{r|}{del : 02}
& \multicolumn{1}{r|}{cls : 03}
& \multicolumn{1}{r|}{mask : 04}
& \multicolumn{1}{r|}{A : 05}
& \multicolumn{1}{r|}{G : 06}
& \multicolumn{1}{r|}{C : 07}
\\ \hline
\multicolumn{1}{|r|}{T : 08}
& \multicolumn{1}{|r|}{GI : 09}
& \multicolumn{1}{r|}{CI : 10}
& \multicolumn{1}{r|}{TI : 11}
& \multicolumn{1}{r|}{AI : 12} 
& \multicolumn{1}{r|}{A,G : 13}
& \multicolumn{1}{r|}{A,C : 14}
& \multicolumn{1}{r|}{G,A : 15}
\\ \hline
\multicolumn{1}{|r|}{G,C : 16}
& \multicolumn{1}{r|}{C,G : 17}
& \multicolumn{1}{|r|}{C,T : 18}
& \multicolumn{1}{r|}{T,C : 19}
& \multicolumn{1}{r|}{GI,del : 20}
& \multicolumn{1}{r|}{T,G : 21}
& \multicolumn{1}{r|}{G,T : 22}
& \multicolumn{1}{r|}{C,A : 23}
\\ \hline
\multicolumn{1}{|r|}{C,ins : 24}
& \multicolumn{1}{r|}{CI,del : 25}
& \multicolumn{1}{r|}{T,ins : 26}
& \multicolumn{1}{|r|}{TI,del : 27}
& \multicolumn{1}{r|}{A,T : 28}
& \multicolumn{1}{r|}{G,ins : 29}
& \multicolumn{1}{r|}{A,ins : 30}
& \multicolumn{1}{r|}{T,A : 31}
\\ \hline
\multicolumn{1}{|r|}{AI,del : 32}
& \multicolumn{7}{r|}{}
\\ \hline
\end{tabular}
\label{tab:encode}
\end{table*}

\subsection*{Model Architecture}

We adopted and substantially extended a transformer-based model architecture originally proposed by Elmes et al. \cite{elmes2022snvformer} for single-nucleotide variant (SNV) analysis for the prediction of gout risk.
Our implementation was specifically designed for LC prediction using only genotype data -- eschewing conventional clinical covariates such as age, sex, and other phenotypic features to isolate the predictive capacity of genomic variation.
Several key architectural modifications were implemented to optimize performance for LC prediction.
We replaced the concatenation-based positional encoding mechanism with additive positional encoding, aligning the model architecture with canonical Transformer designs and improving both learning dynamics and computational efficiency.
A systematic evaluation of the model complexity provided empirical evidence for the optimal number of transformer layers and attention heads required for effective performance.
Most importantly, to support model transparency and biological insight, we incorporated a dedicated explainability module, enabling the identification and interpretation of genomic variants most relevant to LC prediction.

Our transformer model consists of an embedding layer, an encoder, and a classification layer.
Each SNP is represented by a learnable embedding vector instead of one-hot encoding for richer SNP representations. We employed PyTorch’s nn.Embedding layer to map each SNP, encoded as a unique integer index, to a dense vector in a continuous space. The embedding matrix, with shape $N \times d$, where $N$ is the number of unique SNPs and $d$ is the embedding dimension, is randomly initialized and trained jointly with the model. This approach allows the model to learn task-specific representations of genetic variation. To incorporate information about the order of SNPs — which is critical for capturing the sequential structure of genomic data — we added a fixed positional encoding to the SNP embeddings. These encodings are computed using sinusoidal functions of varying frequencies, following the formulation introduced in the Transformer architecture \cite{vaswani2017attention}, enabling the model to leverage relative and absolute positional information without introducing additional trainable parameters.

For the encoder, we used the Linformer architecture \cite{wang2020linformer}, a low-rank approximation of the standard Transformer self-attention mechanism, which reduces the quadratic complexity of attention computation to linear with respect to the sequence length. This is especially beneficial in genomic contexts, where input sequences (i.e., SNP arrays) can be long and computational efficiency becomes critical. By projecting key and value matrices into a lower-dimensional space, Linformer enables efficient modeling of long-range dependencies while significantly reducing memory and computational overhead. This trade-off makes Linformer a practical and scalable choice for genome-wide data analysis compared to standard Transformer models, which are often infeasible for long genomic sequences due to their high computational cost.
The classification head is implemented as a single fully connected (linear) layer.
The end-to-end modeling pipeline is summarized in Fig. \ref{fig:pipeline}.

\subsection*{Training Setup}
Each input SNP token was mapped to a 36-dimensional learnable embedding vector, resulting in an embedding matrix of dimensions $33\times36$, where $33$ represents the total number of unique SNP tokens in our vocabulary.
The embedding dimension of 36 was selected to be slightly larger than the token vocabulary size (33), to allow for expressive, task-specific representations while avoiding over-parameterization that could lead to overfitting. 

The encoder model architecture is composed of a single lightweight transformer encoder block with one attention head.
Given the small embedding size and limited input complexity, we opted for a single attention head in our Transformer encoder.
This choice maximized the per-head representational capacity and avoided the redundancy that often arises in multi-head settings with low-dimensional inputs \cite{michel2019sixteen, voita2019analyzing}.
Our empirical ablation confirmed that increasing the number of heads or layers did not improve model performance.
The Linformer layer in the encoder model used a projection dimension ($k$) of $36$, enabling a linear approximation of the self-attention mechanism.

For primary experiments, we used a 70-10-20 train-validation-test split on the dataset.
The model was trained in two stages: (i) pretraining the encoder using a masked language modeling (MLM) objective, and (ii) fine-tuning with a binary classification head for LC prediction. During pretraining, $40\%$ of the input tokens were randomly selected for masking. Of these, $80\%$ were replaced with a special [MASK] token, $10\%$ were replaced with a random token, and the remaining $10\%$ were left unchanged, following the masking strategy introduced in BERT \cite{devlin2019bert}. The MLM objective encouraged the model to learn contextual representations of genomic variants.

Additionally, we employed five-fold cross-validation to assess model stability across different data partitions.
The full dataset was randomly partitioned into 5 equally sized folds.
For each iteration, one fold was used as the validation set while the remaining four folds were used for training.
This process was repeated 5 times, ensuring that each data point was used for validation exactly once.
The model was reinitialized at the start of each fold, and performance metrics were averaged over all folds to provide a comprehensive evaluation.

We performed fine-tuning using the AMSGrad variant of the AdamW optimizer \cite{loshchilov2019decoupled}, with a learning rate of $10^{-7}$.
We used a batch size of 32 and the cross-entropy loss for model optimization.
To prevent overfitting and reduce training time, we implemented early stopping based on the validation loss with a patience of 5 epochs and a minimum improvement threshold of $10^{-4}$, up to a maximum of 50 training epochs.
This technique helped avoid excessive training on noisy or uninformative signals that could degrade generalization.

Training and inference were conducted on an NVIDIA A100-SXM4-80GB GPU (80 GB VRAM). The system featured two AMD EPYC 7742 CPUs, each with 64 cores, totaling 128 physical cores. The machine had 2.0 TB of RAM, which was critical for efficient data preprocessing and in-memory dataset handling, and large-scale model shuffling.
Despite the availability of large GPU memory, CPU memory played a crucial role, especially in handling memory-intensive operations during data loading and preprocessing.

Model performance during fine-tuning and inference was evaluated using standard binary classification metrics, including Area Under the Receiver Operating Characteristic Curve (AUC), precision, recall, and F1-score.
To determine the optimal threshold for computing precision, recall, and F1-score, we employed Youden’s J statistic on the validation set to maximize the trade-off between sensitivity and specificity.

\subsection*{Explainability Framework}

To gain insight into the contribution of individual genomic variants to model predictions, we implemented an explainability pipeline using Layer Integrated Gradients (LIG) \cite{sundararajan2017axiomatic} from the Captum \cite{kokhlikyan2020captum} library.
LIG quantifies feature importance by computing how the model output changes as the input transitions from a baseline (reference) input to the actual input. 
Specifically, it integrates the gradients of the model output with respect to the input embeddings along this continuous path.
Since the exact integral is often intractable, it is approximated using a Riemann sum over $m$ steps as follows:

$$
\mathrm{LIG}_i(x) \approx (x_i - x'_i) \cdot \frac{1}{m} \sum_{k=1}^m \frac{\partial F\left(L\left(x' + \frac{k}{m}(x - x')\right)\right)}{\partial L_i}
$$

Where:
$x$ is input,
$x'$ is the reference or baseline,
$L(.)$ is the embedding layer output,
$F(.)$ is the output of the model given embedded input, and
$\frac{\partial F}{\partial L_i}$ is the gradient with respect to the embedding of token.

\textit{Model Architecture Context:}
Our model is a transformer-based sequence classifier, where each input sequence represents a fixed-length segment of SNPs encoded and embedded into a continuous space. The model outputs a probability distribution over the two classes (LC vs. control). We used the pre-softmax logits for explainability.

\textit{Objective of Attribution:}
We sought to understand which SNPs within the input sequence are most influential in driving the model’s prediction towards the LC class (class = 1). Attributions were therefore computed with respect to class 1 throughout the explainability analysis.
Because each input sequence is composed of SNPs ordered by chromosome and position, we retain this ordering throughout the attribution analysis to preserve genomic context. This layout also enables downstream visualization using a Manhattan-style plot.

\textit{Reference Input for Integrated Gradients:}
Integrated Gradients (IG) requires a baseline or reference input that represents an ``absence of signal''. The choice of this reference is critical, as it defines the path over which gradients are integrated.
We used the mean embedding vector across all samples in the training set as the reference input for IG.
This reflects a ``typical'' genomic sequence in the cohort.
This follows common practice in transformer-based models. where mean embedding provides smoother and more realistic gradients compared to the zero embedding \cite{atanasova2024diagnostic, sanyalren2021discretized}.
Formally, given embeddings $E_i \in \mathbb{R}^{L \times D}$ for all $i \in \{1,...,N\}$, we computed:

  $$
  E_{\text{mean}} = \frac{1}{N} \sum_{i=1}^{N} E_i
  $$

Here, L is the sequence length, and D is the embedding dimension. 
This averaged embedding was used as the baseline reference for LIG computations.

\textit{Target Class Selection:}
To isolate attribution signals specific to LC risk, we computed gradients with respect to the class 1 output.
For model with output $F(x) \in \mathbb{R}^2$, we set `target=1' in the LIG function to extract class-specific attributions.
This ensures the attributions represent genomic features pushing the prediction toward LC (positive) or away (negative).

\textit{Layer Selection for Feature Attribution:}
We targeted the embedding layer immediately prior to Transformer encoding to capture raw genomic signals before contextualization through self-attention mechanisms.
This layer choice, implemented via Captum's `LayerIntegratedGradients' allows direct attribution to individual SNP tokens while avoiding potential confounding from cross-variant interactions introduced by the attention mechanism.

\textit{Computation of Attributions:}
For each patient in the test set, we computed LIG using the selected reference and class 1 as the target output.
This process generated attribution tensors representing importance scores for each SNP position across all embedding dimensions.
We then aggregated these attribution scores by computing the mean across the embedding dimensions, which provided a single attribution value per SNP position.
This process was repeated for all test samples.

\textit{Aggregation Across Individuals:}
To quantify the predictive effect of each genomic variant (chromosome, position, SNP token) tuple, we computed average attribution scores across only for those individuals carrying the variant of interest.
This conditional aggregation strategy highlights variant-specific effects while avoiding signal dilution from non-carriers, where the variant has no influence on disease risk.

\textit{Cross-Validation Attribution:}
To ensure robustness, we repeated the attribution process for each of the 5 cross-validation folds.
For each fold, we trained a model and computed Layer IG on the test set of that fold using the mean embedding baseline from the training set of the same fold.
This ensured that no test sample contributes to the baseline embedding of its own fold, preserving separation.
We then aggregated attribution scores per (chromosome, position, SNP token) tuple across folds by averaging test-set attributions across all folds.
This multi-fold approach reduces variance from random train/test data partitioning and provides more reliable identification of consistently important genomic variants.

\textit{Visualization: Manhattan Plot Analogy:}
We adapted the concept of a Manhattan plot to visualize SNP importance.
In our Manhattan-style attribution plot, the x-axis is constructed by concatenating all chromosomes sequentially, in numerical order, so that SNPs from chromosome 1 occupy the first section of the x-axis, followed by SNPs from chromosome 2, and so on, up to the last chromosome included in the input.
This layout mirrors the traditional genome-wide Manhattan plot in GWAS, enabling visual identification of chromosome-specific attribution peaks.
Peaks in this plot correspond to regions where the model assigns high importance (positive or negative) to SNPs for the prediction task (e.g., LC classification).
To improve readability chromosome boundaries are marked on the x-axis with alternating colors.
The y-axis represents the mean attribution scores.
Note that although this plot reflects attribution scores (not p-values), its structure is inspired by GWAS plots, emphasizing interpretable alignment with genomic architecture.
Peaks in the plot indicate SNPs with consistently high attribution toward LC predictions.
This visualization helps reveal the SNPs that drive the model decisions and may correspond to biologically meaningful variants.
Positive attributions indicates SNPs that support the model confidence in predicting LC.
Negative attributions indicates SNPs that diminish the LC prediction, potentially protective or neutral.
We focused primarily on attributions toward class 1, but this framework is extensible to class 0 as well.

\section*{Results}

In this section, we present the results from our experiments, including model architecture selection through ablation studies, evaluation of the final architecture using cross-validation, and explainability analysis to understand the model's behavior and biological relevance.

\subsection*{Ablation Study}
Our experimental procedure was organized to incrementally refine the model architecture and assess its performance.
We began with a 70-10-20 train-validation-test split to conduct model architecture selection and data filtering (MAF thresholding).
This fixed split allowed consistent comparison during initial ablation studies.

\subsubsection*{Model architecture}
\begin{figure}[!htbp]
    \centering
    \includegraphics[width=\linewidth]{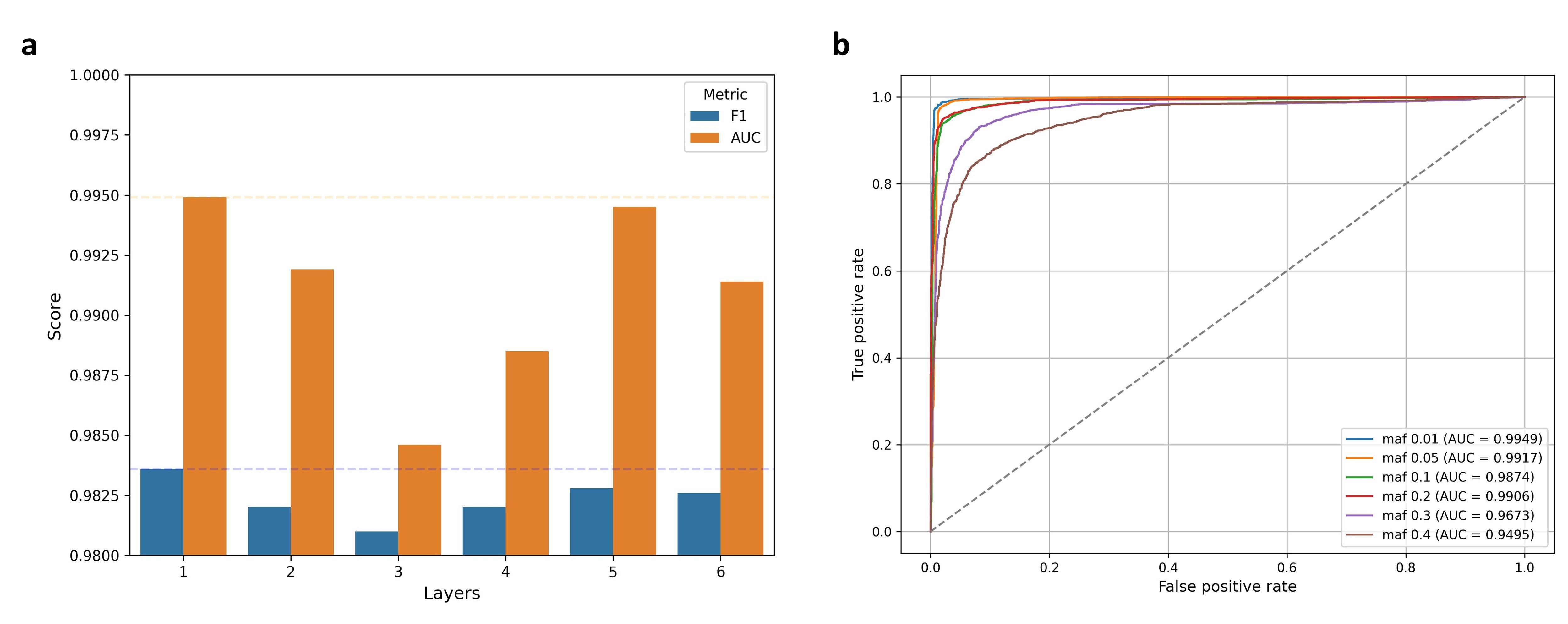}
    \caption{Ablation study with varying transformer depth and minor allele frequency (MAF) threshold. \textbf{a}, Effect of transformer depth on model performance, assessed by varying the number of encoder layers from 1 to 6 while keeping the number of attention heads fixed at 1. \textbf{b}, Effect of MAF threshold on model performance, assessed by varying the MAF thresholds 0.01, 0.05, 0.1, 0.2, 0.3, and 0.4.}
    \label{fig:abl_layer_maf}
\end{figure}

We first examined how the complexity of the transformer architecture influenced performance. 
Specifically, we varied the number of encoder layers (1–6) while keeping the number of attention heads fixed at 1.
All experiments used the fixed 70-10-20 split.
The AUC and F1 scores presented in Fig. \ref{fig:abl_layer_maf}a showed that increasing depth beyond a single layer offered no consistent improvement in validation performance. In fact, deeper models occasionally showed greater variance, suggesting overfitting. Based on this, we selected 1 layer as the optimal depth for further experiments.

We also tested different numbers of attention heads while holding the number of transformer layers fixed at 1. Despite the long input sequences (378,866 positions), we found that increasing the number of heads beyond one did not yield any improvements in model performance. This may be attributed to the structured and sparse nature of SNP data, where most positions are uninformative and the important signals are relatively localized. Additionally, the combination of a low-dimensional embedding space (36) and a modest vocabulary size (33 SNP tokens) may have limited the benefits of multi-head attention. Based on these findings, we adopted a simple and efficient architecture with a single attention head and one transformer layer, which offered both strong performance and reduced computational complexity.

\subsubsection*{Minor Allele Frequency (MAF) Thresholding}

We next investigated the impact of filtering genetic variants based on MAF, using the same 70-10-20 data split.
We evaluated multiple MAF thresholds -- 0.01, 0.05, 0.1, 0.2, 0.3, and 0.4 -- to examine how the inclusion or exclusion of less frequent variants affected predictive performance.
As shown in Fig. \ref{fig:abl_layer_maf}b, we observed that AUC score declined as the MAF threshold increased beyond 0.01, indicating that excluding less frequent variants led to a loss of predictive signal.
This suggests that consistent with our previous studies, rare variants -- despite their low frequency -- carry important information relevant to lung cancer prediction \cite{selvan2020inherited, esai2019rare}. Based on these results, we selected the MAF threshold that yielded the highest performance for use in subsequent analyses, which was 0.01.

\subsection*{5-fold Cross-validation}

To obtain a robust estimate of model generalization and reduce potential biases arising from a single data split, we employed 5-fold cross-validation following the ablation study.
Cross-validation systematically partitions the data into multiple train-validation splits, ensuring that each sample is used for both training and validation exactly once across folds.
This procedure mitigated variance in performance estimates, allowed for a more comprehensive evaluation of the model stability, and reduced the risk of overfitting to a particular subset of the data.

\begin{figure}[!htbp]
    \centering
    \includegraphics[width=0.7\linewidth]{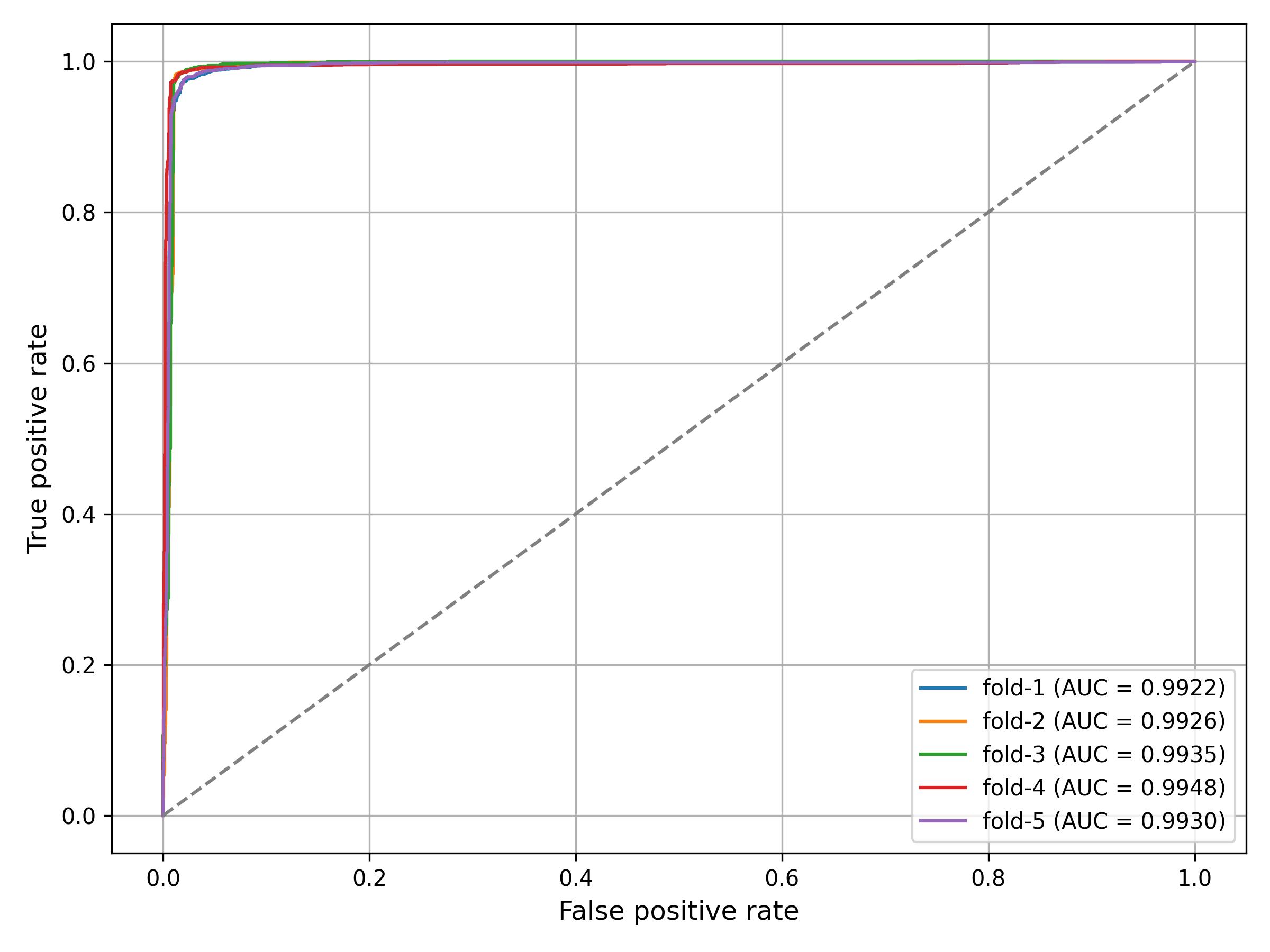}
    \caption{Model performance across 5-fold cross-validation.}
    \label{fig:cv}
\end{figure}

Fig. \ref{fig:cv} illustrates the AUC scores across all cross-validation folds, and Table \ref{tab:cv} summarizes precision, recall, F1, and AUC scores for each fold.
The table also includes the average scores with standard deviation and the corresponding 95\% confidence intervals across folds, providing a comprehensive view of model stability and generalization.
The model achieved an average AUC of 0.9932$\pm$0.0010, with consistently high precision, recall, and F1 scores across all folds.
These results confirmed that the model generalizes well across data partitions and is not overfitting to any particular subset.

\begin{table}[!htbp]
\caption{Cross-validation performance across 5 folds.}
\centering
\begin{tabular}{|l|c|c|c|c|c|c|c|}
\hline
\multicolumn{1}{|c|}{Metric} &
\multicolumn{1}{|c|}{Fold 1} & 
\multicolumn{1}{|c|}{Fold 2} & 
\multicolumn{1}{|c|}{Fold 3} & 
\multicolumn{1}{|c|}{Fold 4} & 
\multicolumn{1}{|c|}{Fold 5} & 
\multicolumn{1}{|c|}{Mean$\pm$STD} & 
\multicolumn{1}{|c|}{95\% CI} \\ \hline
Precision              & 0.9864                 & 0.9858                 & 0.9813                 & 0.9851                 & 0.9743                 & 0.9826$\pm$0.0050                 & [0.9686,0.9966]          \\ \hline
Recall                 & 0.9732                 & 0.9835                 & 0.9844                 & 0.9829                 & 0.9793                 & 0.9807$\pm$0.0046                 & [0.9679,0.9934]          \\ \hline
F1-score               & 0.9798                 & 0.9846                 & 0.9828                 & 0.984                  & 0.9768                 & 0.9816$\pm$0.0033                 & [0.9726,0.9906]          \\ \hline
AUC                    & 0.9922                 & 0.9926                 & 0.9935                 & 0.9948                 & 0.9930                  & 0.9932$\pm$0.0010                 & [0.9904,0.9960]          \\ \hline
\end{tabular}
\label{tab:cv}
\end{table}

\subsection*{Explainability Analysis}
To identify genomic variants that contributed most strongly to the model predictions, we applied Layer Integrated Gradients to the trained models from each cross-validation fold. Attributions were computed with respect to the class 1 (LC) output, using the mean embedding vector from the training set of each fold as the reference baseline. We attributed importance at the embedding layer and aggregated results across embedding dimensions, samples, and cross-validation folds.

\begin{figure}[!htbp]
\centering
\includegraphics[width=\linewidth]{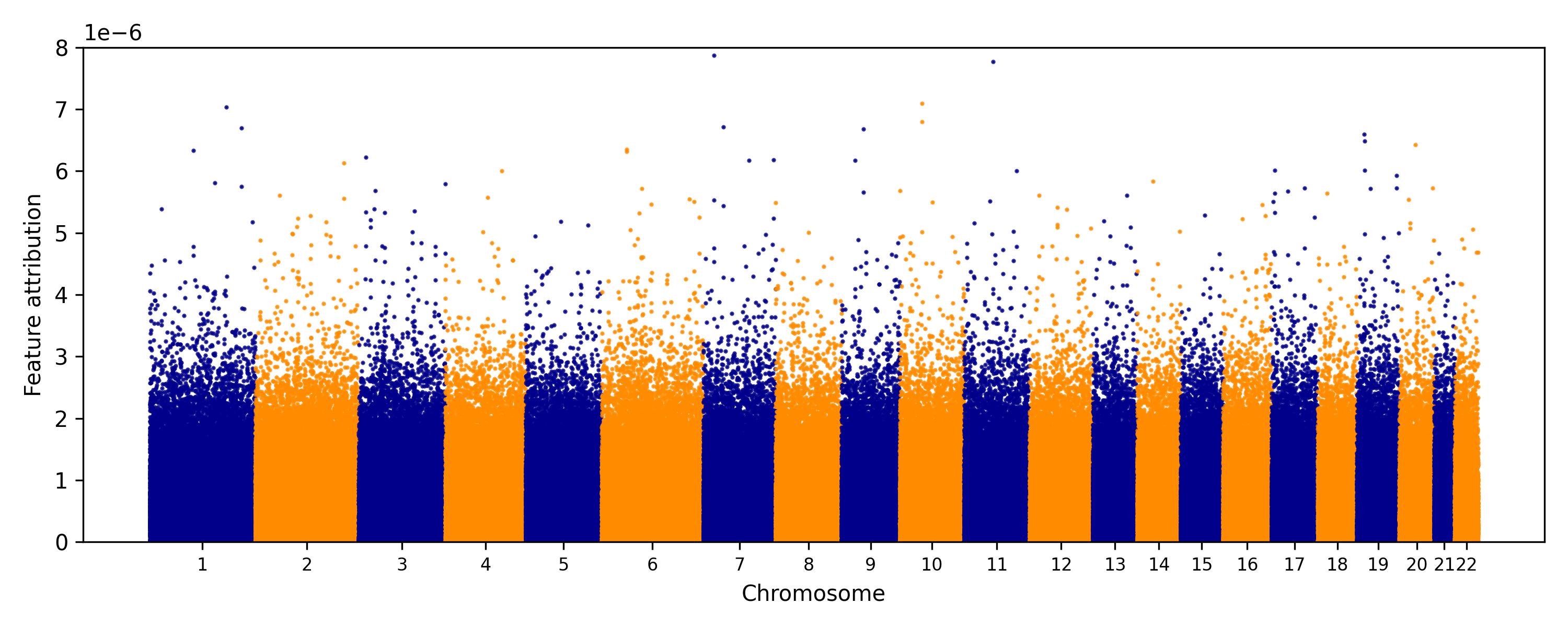}
\caption{Manhattan-style plot of SNP attribution scores across the genome.
Each point represents a single nucleotide polymorphism (SNP), with its genomic position on the x-axis and its average positive attribution score (with respect to lung cancer prediction) on the y-axis.
Chromosomes are concatenated end-to-end along the x-axis and alternately colored for visual clarity.
Only SNPs with positive attribution scores are shown, highlighting features that contribute positively to the model’s classification of lung cancer.}
\label{fig:manhattan_plot}
\end{figure}

In Fig. \ref{fig:manhattan_plot}, we present these attribution scores in a Manhattan-style plot, where each point corresponds to a SNP located at a specific base-pair position.
SNPs are grouped and alternately colored by chromosome.
The x-axis spans the genomic positions in a contiguous fashion, with chromosomes aligned end-to-end, while the y-axis displays the average attribution score for each SNP across the test sets.
The figure shows the SNPs predictive of a lung cancer diagnosis with positive attribution scores in model confidence.
In conventional GWAS, association signals often appear as broad peaks in Manhattan plots, where many correlated SNPs share low p-values due to linkage disequilibrium (LD). In contrast, attribution Manhattan plots from our transformer model -- computed with Layer Integrated Gradients -- highlight a more focal set of variants. Rather than distributing importance across an entire LD block, the model tends to assign elevated attribution to a limited subset of SNPs. This pattern suggests that the model represents predictive information differently from statistical association tests. However, concentrated attribution should not be conflated with pinpointing causal variants, as gradient-based explanations reflect the model’s internal reliance on features, not biological causality.

To assess the biological relevance of these attributions, we compared the genomic positions of the top-ranked GWAS SNPs to loci reported in two well-established LC studies \cite{gorman2024multi, lee2021functional}.
Specifically, we selected the 50 most positively attributed SNPs per chromosome.
Given that our dataset contains unimputed SNPs, the variant resolution is inherently sparser compared to those used in large-scale GWAS, making exact positional matches less likely.
To address this, we implemented a $\pm$1 million base-pair (1 Mbp) window-based proximity search.
This biologically informed buffer accounts for potential linkage disequilibrium and different SNPs influencing the same gene through \textit{cis}-regulatory effects \cite{baca2022genetic, davis2016efficient}.

We considered a match if a top-attributed SNP locus in our model appeared within $\pm$1 Mbp of a known lung cancer-associated locus, based on chromosome and base-pair position.
This approach enabled us to validate model-driven signals even when the lead SNP in previously reported lung GWAS were not present on the genotyping array used here.

The genomic positions of several highly ranked SNPs showed strong positional concordance with established lung cancer susceptibility loci, lending support to the biological validity of the model attributions and suggesting that it has captured meaningful genomic patterns.
Table \ref{tab:validated_snps} summarizes these validated SNPs, showing those that fall within $\pm$1Mbp of previously reported LC-associated variants.

Each entry includes the SNP chromosomal location and whether it falls within proximity to loci identified in prior studies.
Because our dataset was unimputed and contained no direct overlaps with known LC susceptibility loci, we did not include rsIDs or allele information.
This table highlights specific examples where the model predictions aligned with known biology, providing confidence in its explainability and potential utility in future genomic research.
The top risk variants on Chromosome 6 listed in Table \ref{tab:validated_snps} span the complete MHC region (chr6: 29.9-33.2 Mb in GRCh37/b37) and encompass classical HLA Class I and II genes, Class III complement and cytokine genes, and extended MHC regulatory elements. These risk variants suggest potential disruption of coordinated antigen presentation, immune tolerance, and inflammatory responses that are critical for tumor immunosurveillance. Additional risk variants at chr6: 10.1 Mb and 26.5-27.3 Mb may further compromise immune function through effects on other chromosome 6 immune-related genes, collectively undermining HLA-mediated pathogen resistance and autoimmune regulation in providing protection against malignant transformation.

\begin{table}[!htbp]
\centering
\caption{Genomic positions of top attributed SNPs for the model’s LC risk predictions that fall within $\pm$1 Mbp of known lung cancer-associated loci based on comparisons with two well-established studies\cite{gorman2024multi, lee2021functional}.}
\begin{tabular}{|C{3.5cm}|C{7cm}|C{5.5cm}|}
\hline
Chromosome & Top attributed SNP positions within $\pm$1 Mbp & Known lung cancer loci \\ \hline
1 & 160386089 & 160210727\cite{lee2021functional}  \\ \hline
5 & 133223816 & 133864599\cite{gorman2024multi} \\ \hline
6 & 29910698, 30340145, 30721933, 31324615, 31369151, 31638848 & 30882415\cite{lee2021functional} \\ \hline
6 & 31638848, 32017521, 32025870, 32292956, 32513102, 32608537, 32687973, 32796019, 32822186, 33037419, 33192867 & 32591476\cite{lee2021functional}, 32605884\cite{lee2021functional} \\ \hline
6 & 10114925 & 10415006\cite{lee2021functional} \\ \hline
6 & 26459997, 27279877 & 26328353\cite{lee2021functional}, 26403036\cite{lee2021functional}, 26581258\cite{lee2021functional}, 26651053\cite{lee2021functional}, 26686131\cite{lee2021functional} \\ \hline
8 & 128885474, 129299946 & 129535264\cite{gorman2024multi} \\ \hline
10 & 4442508 & 4961021\cite{lee2021functional} \\ \hline
11 & 126355993 & 125510257\cite{lee2021functional} \\ \hline
12 & 48526711 & 47857826\cite{gorman2024multi} \\ \hline
12 & 8450495 & 9058562\cite{lee2021functional} \\ \hline
12 & 127711416 & 127225803\cite{gorman2024multi} \\ \hline
15 & 70634116, 70734621, 70770766 & 70431773\cite{gorman2024multi} \\ \hline
19 & 16860558, 17212992 & 17401859\cite{gorman2024multi} \\ \hline
21 & 39815520 & 40173528\cite{lee2021functional} \\ \hline
\end{tabular}
\label{tab:validated_snps}\end{table}

\begin{figure}[!htbp]
\centering
\includegraphics[width=\linewidth]{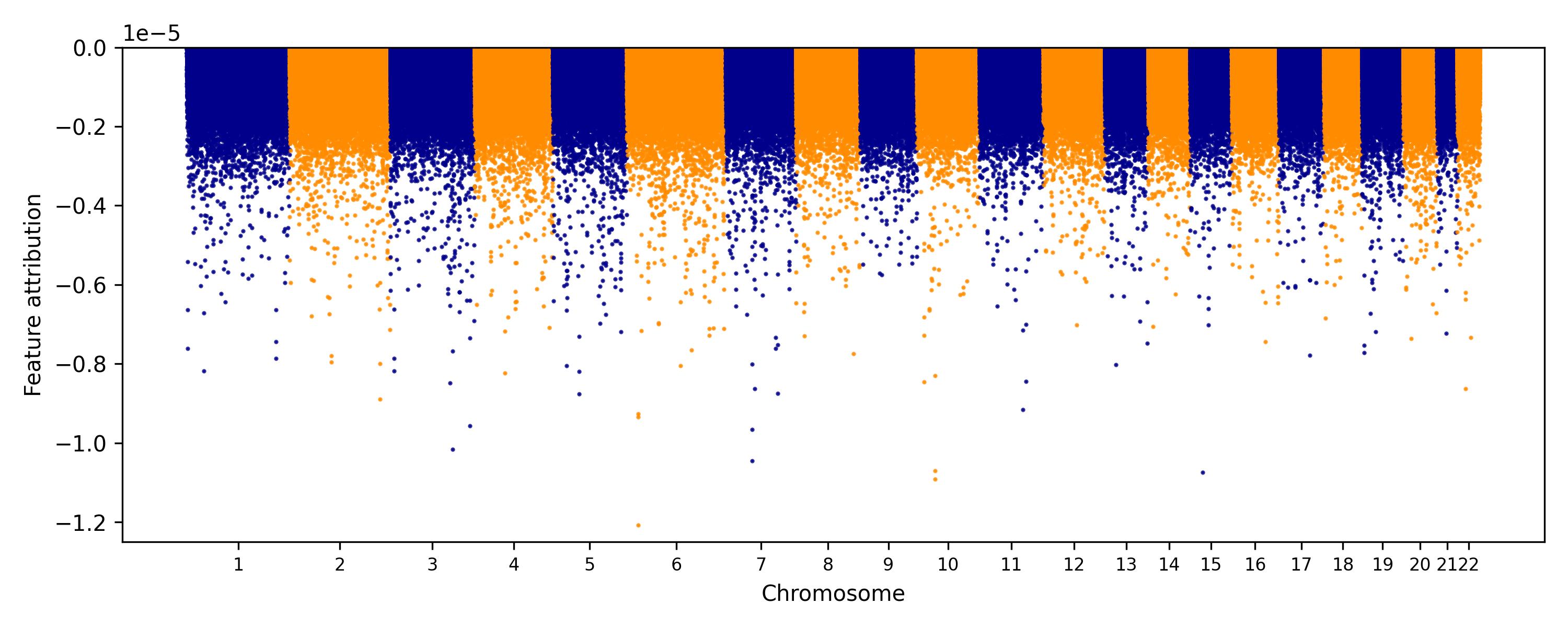}
\caption{Manhattan-style plot of negative SNP attribution scores across the genome.
Each point represents a single nucleotide polymorphism (SNP), with its genomic position on the x-axis and its average negative attribution score (with respect to lung cancer prediction) on the y-axis.
Chromosomes are concatenated end-to-end along the x-axis and alternately colored for visual clarity.
Only SNPs with negative attribution scores are shown, highlighting features that contribute negatively to the model’s classification of lung cancer.}
\label{fig:manhattan_plot_neg}
\end{figure}

Additionally, we explored negative attribution scores as shown in Fig. \ref{fig:manhattan_plot_neg}, which may correspond to protective variants.
We identified the following five most putative protective loci: chr6:19841493, chr10:31409908, chr15:46320085, chr7:50173777, and chr3:148789127, ordered by ascending attribution scores. 
Notably, the locus with the strongest negative signal, chr6:19841493, lies in the telomeric region of chromosome 6p, approximately 10Mb upstream of the canonical MHC region (chr6: \textasciitilde29.5–33.4 Mb), which is known to harbor immune-related genes \cite{zeynep2024animmunogenetic}.
Although this variant does not fall within the classical MHC locus, given our recent work that showed Human Leukociyte Antigen (HLA) class II heterozygosity is associated with lower LC risk \cite{krishna2024immunogenetic}, its telomeric location on chromosome 6p suggests potential immune regulatory involvement through long-range interactions or shared regulatory networks.

The remaining negatively attributed loci or their proximal genes do not map to known LC susceptibility loci.
This suggests the possibility that negative attributions may reflect interactions or compensatory mechanisms rather than simple protective alleles.
For example, a variant that buffers the effect of a nearby risk variant (epistatic interaction) or one that modulates gene expression in a tissue-specific way could still receive a negative score in the model.
However, functional validation through further experimental or computational analyses is necessary to elucidate their roles.

\section*{Discussion}

This study demonstrates that transformer-based models, when tailored appropriately, can effectively capture predictive genomic signals for LC classification using raw SNP sequences.
Through a series of ablation experiments, we found that a lightweight architecture with a single attention head and transformer layer is sufficient for this task.
Despite the high dimensionality and sparsity of genomic data, our model achieved robust performance across all folds in a 5-fold cross-validation, with an average AUC exceeding 0.99.
We also investigated the impact of variant filtering through minor allele frequency (MAF) thresholds and observed that retaining low-frequency variants improved predictive performance.
This highlights the utility of incorporating less frequent variants in genetic disease models, even when using unimputed genotype data.
Our model demonstrates that competitive prediction performance is achievable using only raw genotype data alone, without reliance on traditional clinical variables, highlighting the power of deep genomic representations for complex disease risk modeling in precision medicine applications.

Beyond performance metrics, we conducted a comprehensive explainability analysis using Layer Integrated Gradients (LIG). By attributing importance scores to input SNPs, we identified features that most strongly influenced the model predictions for the LC class. Importantly, many of the top-scoring features aligned with known susceptibility loci reported in large-scale GWAS and functional annotation studies \cite{gorman2024multi, lee2021functional}. We employed a $\pm$1 million base pair window to account for discrepancies due to unimputed variant representation. Our explainability framework also revealed negatively attributed SNPs—variants that potentially push the model away from predicting lung cancer. While some of these did not directly overlap with known protective loci, they may correspond to regulatory or epigenetic mechanisms yet to be fully characterized. This warrants further exploration.

Twin based studies had estimated the heritability of LC to be around 18\% \cite{mucci2016familial}; this suggests that the maximum achievable AUC would be around 0.65, much lower than what we observed here \cite{wray2010genetic}.
As we evaluated the AUC in a strictly held-out test set of individuals, we do not believe that these high AUCs are due to overtraining on the same samples we evaluated.
Instead, there are several possible explanations.
First, much of the theoretical work on the relationship between polygenic risk prediction and AUC assumes additive effects under a liability threshold model \cite{klein2022polygenic}; dominance and epistatic effects are also captured by our deep learning approach.
Second, the SNP data was generated on DNA isolated from blood from adults, some of whom already may have either diagnosed or undiagnosed lung cancer.
Changes in the copy number of genomic segments of DNA in the blood could alter the ability of the genotyping algorithm to call a genotype, producing ``nan'' tokens in our model.
Thus, it is theoretically possible for our approach to include information about cancer status if there is a signal to be found in circulating blood DNA \cite{tian2023clonal}.
Along those lines, we note that clonal mutations in hematopoietic stem and progenitor cells have been associated with LC risk \cite{bauml2018clonal, esai2025distinct}.
We recognize that further validation of this model in independent datasets is necessary; data security restrictions prevented us from moving this model to be able to be used on other datasets.

\textbf{Limitations:}

Despite these promising findings, there are several limitations to this study.
Unimputed genotype data limited our ability to capture the full spectrum of genomic variation and may have affected resolution when matching to GWAS findings.
Using whole genome sequencing data could reveal additional relevant variants, particularly in non-coding regions.
Explainability is inherently approximate.
While Layer Integrated Gradients provides insight into feature relevance, attribution scores depend on the choice of reference, embedding structure, and model non-linearities.
Biological interpretations should be treated as hypothesis-generating rather than conclusive and evaluation of negative attributions is limited.
Our exploration of negatively scoring SNPs is preliminary and requires deeper biological modeling to determine the mechanism of action.

This work presents a simple yet powerful transformer-based model for genomic sequence classification and demonstrates how attribution-based explainability can bridge predictive performance with biological relevance.
Our approach effectively prioritizes putative risk variants and validates them against independent literature, despite working with raw unimputed SNP data.
By leveraging model transparency, we move beyond “black-box” prediction and contribute toward a more explainable and hypothesis-driven application of deep learning in genomics.
Future research should extend this framework to larger, more diverse cohorts, incorporate imputed data, and integrate multi-omic signals to enhance both predictive and biological resolution.


\section*{Acknowledgements}
This work is sponsored by the US Department of Veterans Affairs using resources from the Knowledge Discovery Infrastructure which is located at the Oak Ridge National Laboratory and supported by the Office of Science of the U.S. Department of Energy. This manuscript has been authored by UT-Battelle, LLC, under contract DE-AC05-00OR22725 with the US Department of Energy (DOE). The US government retains and the publisher, by accepting the article for publication, acknowledges that the US government retains a nonexclusive, paid-up, irrevocable, worldwide license to publish or reproduce the published form of this manuscript or allow others to do so for US government purposes. DOE will provide public access to these results of federally sponsored research in accordance with the DOE Public Access Plan (\url{http://energy.gov/downloads/doe-public-access-plan}).

The authors would like to thank Mrs. Hope Cook for her guidance with the data query optimization.

\section*{Funding}
This research is based on data from the Million Veteran Program, Office of Research and Development, Veterans Health Administration, and was supported by MVP000 as well as award MVP061. This publication does not represent the views of the Department of Veteran Affairs or the United States Government. The authors also wish to acknowledge the support of the larger DOE-VA partnership. Most importantly, the authors would like to thank and acknowledge the veterans who chose to get their care at the VA.

\section*{Data availability}
The dataset developed for this study is not accessible to the public under requirements of the Health Insurance Portability and Accountability Act of 1996 and related privacy and security concerns. The data underlying this publication are accessible to researchers with Million Veteran Program (MVP) data access. MVP is currently only accessible to researchers who have a funded MVP project.

\section*{Code availability}
The code for data preprocessing, model training, and performance evaluation is available on GitHub at: \url{https://github.com/mmahbub/HEMERA}.

\section*{Author contributions statement}

M.M. and I.D. conceptualized the study.
M.M. designed the study, developed the study pipeline and software, preprocessed data, performed visualization, and prepared the manuscript with input from all authors.
I.D. and I.G. curated the cohort data.
M.M., I.D., R.K., M.E.S., and Z.H.G. performed the formal analysis of the results.
R.K., M.E.S., and Z.H.G. provided feedback throughout the study to guide the experiments and analysis.
S.A., I.D., and Z.H.G. acquired funding for the project.
All authors reviewed the manuscript and provided feedback.


\end{document}